\documentclass{article}

\usepackage[preprint, nonatbib]{neurips_2020}

\usepackage[utf8]{inputenc} 
\usepackage[T1]{fontenc}    
\usepackage{hyperref}       
\usepackage{url}            
\usepackage{booktabs}       
\usepackage{amsfonts}       
\usepackage{nicefrac}       
\usepackage{microtype}      

\usepackage{graphicx}
\usepackage{enumitem}

\title{The Joy of Neural Painting}

\author{%
  Ernesto Diaz-Aviles\thanks{We provide the code of our Neural Painters implementaion and notebooks to reproduce our experiments at: \url{https://github.com/libreai/neural-painters-x}} \\
  Libre AI \\
  \texttt{ernesto@libreai.com} \\
  \And
  Claudia Orellana-Rodriguez \\
  Libre AI \\
  \texttt{claudia@libreai.com} \\
  \And
  Beth Jochim \\
  Libre AI \\
  \texttt{beth@libreai.com} \\
}

\begin{document}

\maketitle

\begin{abstract}
  \emph{Neural Painters} is a class of models that follows a GAN framework to generate brushstrokes, which are then composed to create paintings. GANs are great generative models for AI Art but they are known to be notoriously difficult to train. To overcome GAN's limitations and to speed up the Neural Painter training, we applied Transfer Learning to the process reducing it from days to only hours, while achieving the same level of visual aesthetics in the final paintings generated. We report our approach and results in this work.
\end{abstract}

\section{Introduction}
Artists do not learn how to paint pixel by pixel. However, most of the current generative AI Art methods are still centered to teach machines how to 'paint' at the pixel-level in order to achieve or mimic some painting style, e.g., GANs-based approaches and style transfer. This might be effective, but not very intuitive, specially when explaining this process to artists, who are more familiar with colors and brushstrokes.

\begin{figure}[h!]
  \label{fig:artwork}
  \centering
  \includegraphics[width=\linewidth]{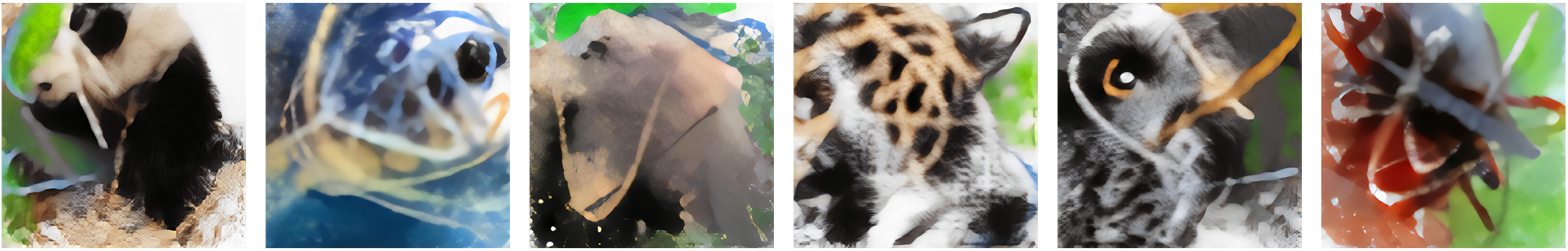}
  \caption{Digital paintings generated using our approach and enlarged with super-resolution~\cite{decrapify}. A combination of brushstrokes help visualize ImageNet classes and can also ``paint'' arbitrary images minimizing a perceptual loss.}
\end{figure}

Part of our goal is to make more accessible the advances of AI to groups of artists who do not necessarily have a deep technical background. We want to explore how the creative process is enriched by the interaction between creative people and creative machines. To this end, we need to teach a machine how to paint as a human would do it: using brushstrokes and combining colors on a canvas (Figure~\ref{fig:artwork}).

\section{Learning Neural Painters Faster}
Neural Painters~\cite{2019NeuralPainters} are a class of models that can be seen as a fully differentiable simulation of a particular non-differentiable painting program, in other words, the machine ``paints'' by successively generating brushstrokes (i.e., actions that define a brushstrokes) and applying them on a canvas, as an artist would do.

These actions characterize the brushstrokes and consist of 12-dimensional vectors defining the following variables: (i)~Start and end pressure; (ii)~Brush size; (iii)~Color (RGB) of the brushstroke; (iv)~Brush coordinates defining the brushstroke’s shape.

The goal of the Neural Painter is to translate these vectors of actions into brushstrokes on a canvas.

One of the Neural Painters architectures is based on GANs. GANs are generative models widely use for AI Art, but they are known to be notoriously difficult to train, specially due to requiring a large amount of data, and therefore, needing large computational power on GPUs. They also require a lot of time to train and are sensitive to small hyperparameter variations.  Reproducing the results of the original Neural Painters paper~\cite{2019NeuralPainters} we obtained similar results with our implementation, in terms of brushstrokes quality, but it took us two days for training to get descent aesthetics using a single GPU on a Colaboratory notebook~\cite{colab}.

To overcome these known GANs limitations, we leverage \emph{Transfer Learning}~\cite{learningToLearn,Goodfellow-et-al-2016} principles that allow us significantly reducing the Neural Painter training time.

Transfer learning is a very useful technique in Machine Learning, e.g., the ImageNet models trained as classifiers, are largely used as powerful image feature extractors, in NLP, word embeddings, learned unsupervised or with minimal supervision, have been very useful as representations of words in more complex language models.

The fundamental idea, is to learn a model or feature representation on a task, and then transfer that knowledge to another related task, without the need to start from scratch, and only do some fine-tuning to adapt the model or representation parameters on that task.

More precisely, since GANs main components are the Generator and Critic the idea is to pre-train them independently, that is in a non-adversarial manner, and do transfer learning by hooking them together after pre-training and proceed with the adversarial training, i.e., GAN mode. This process has shown to produce remarkable results~\cite{decrapify} and is the one we follow here. The main steps are as described as follows:

\begin{enumerate}[wide]
  \item \textbf{Pre-train the Generator with a non-adversarial loss}, e.g., using a feature loss (also known as perceptual loss)~\cite{johnson2016perceptual}

  \item \textbf{Freeze the pre-trained Generator weights}
  
  \item \textbf{Pre-train the Critic as a Binary Classifier} (i.e., non-adversarially) using the pre-trained Generator (in evaluation mode with frozen model weights) to generate fake brushstrokes. That is, the Critic should learn to discriminate between real images and the generated ones. This step uses a standard binary classification loss, i.e., Binary Cross Entropy, not a GAN loss
  
  \item \textbf{Transfer learning for adversarial training (GAN mode)}: continue the Generator and Critic training in a GAN setting (Faster)
  
\end{enumerate}

\paragraph*{From Brushstrokes to Paintings.}Once the Generator training process is completed, we have a machine that is able to translate vectors of actions to brushstrokes, in order to generate a painting we use Intrinsic Style Transfer~\cite{2019NeuralPainters}, similar in spirit to Neural Style Transfer~\cite{gatys2015neural} but which does not require a style image. Intuitively, the features of the content input image and the one produced by the Neural Painter should be similar. To implement the process, we freeze the Generator model weights and learn a set of action vectors that when input to the Generator will produce brushstrokes, that once combined, will create a painting given an input content image. This process can be also used to visualize ImageNet classes. The image features are extracted using a VGG16~\cite{simonyan2015deep} network as a feature extractor.

\section{Conclusion}
In this work we have combined the benefits of Neural Painters and Transfer Learning to reduce the model training time, while retaining the quality aesthetics for the brushstrokes and paintings generated. Our contribution allows practitioner artists to experiment with more models that require hours and not days to train. We are currently working on an approach that does not require a GAN framework, i.e., using only non-adversarial perceptual losses, to further reduce training time and computational resources for artwork generation.

\paragraph*{Acknowledgments.} We would like to thank Reiichiro Nakano for helping us clarifying doubts during the implementation of our Neural Painters and for his supportive and encouraging comments and feedback.

\bibliographystyle{plain}
\bibliography{biblio}

\begin{thebibliography}{1}

\bibitem{decrapify}
Jason Antic, Jeremy Howard, and Uri Manor.
\newblock Decrappification, deoldification, and super resolution.
\newblock \url{https://www.fast.ai/2019/05/03/decrappify/}, 2019.

\bibitem{gatys2015neural}
Leon~A. Gatys, Alexander~S. Ecker, and Matthias Bethge.
\newblock A neural algorithm of artistic style, 2015.

\bibitem{Goodfellow-et-al-2016}
Ian Goodfellow, Yoshua Bengio, and Aaron Courville.
\newblock {\em Deep Learning}.
\newblock MIT Press, 2016.
\newblock \url{http://www.deeplearningbook.org}.

\bibitem{johnson2016perceptual}
Justin Johnson, Alexandre Alahi, and Li~Fei-Fei.
\newblock Perceptual losses for real-time style transfer and super-resolution,
  2016.

\bibitem{2019NeuralPainters}
Reiichiro Nakano.
\newblock Neural painters: {A} learned differentiable constraint for generating
  brushstroke paintings.
\newblock {\em CoRR}, abs/1904.08410, 2019.

\bibitem{colab}
Google Research.
\newblock Collaboratory.
\newblock \url{https://colab.research.google.com/}, 2020.

\bibitem{simonyan2015deep}
Karen Simonyan and Andrew Zisserman.
\newblock Very deep convolutional networks for large-scale image recognition,
  2015.

\bibitem{learningToLearn}
S.~Thrun and L.~Pratt.
\newblock {\em Learning to Learn}.
\newblock Springer US, 2012.

\end{thebibliography}

\appendix
\section*{Supplementary Material}

\section{Learning Neural Painters Faster. Details.}
\subsection{Pre-train the Generator with a Non-Adversarial Loss}

\begin{figure}[ht!]
  \label{fig:pre-train-generator}
  \centering
  \includegraphics[width=0.5\linewidth]{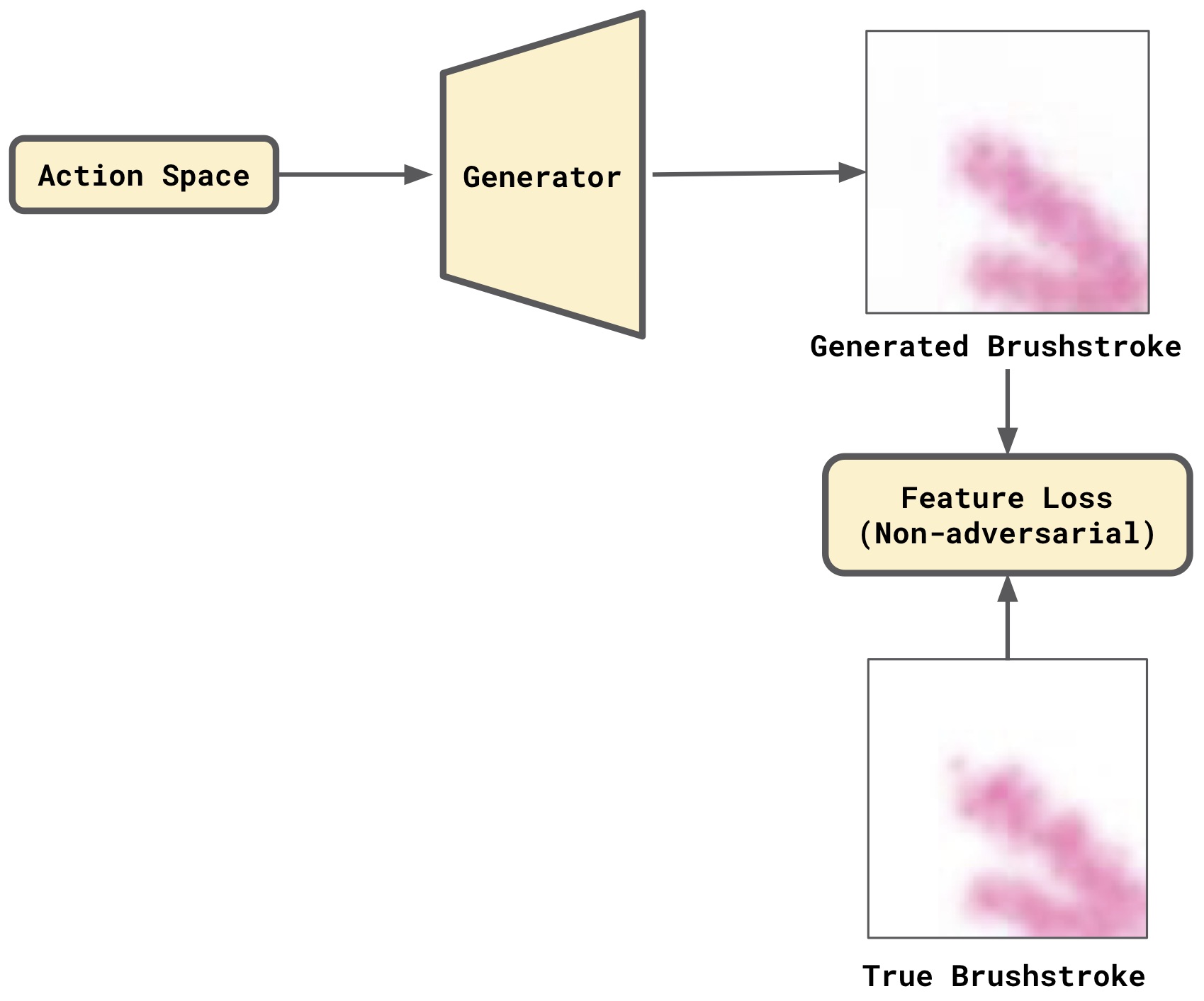}
  \caption{Pre-train the Generator using a (non-adversarial) feature loss.}
\end{figure}

The training set consists of labeled examples where the input corresponds to an action vector and the corresponding brushstroke image to the target.  The input action vectors go through the Generator, which consists of a fully-connected layer (to increase the input dimensions) and of a Deep Convolutional Neural Network connected to it. The output of the Generator is an image of a brushstroke. The loss computed between the images is the feature loss or perceptual loss~\cite{johnson2016perceptual}. The process is depicted in Figure~\ref{fig:pre-train-generator}

\subsection{Freeze the pre-trained Generator}
\begin{figure}[ht!]
  \label{fig:sample-brushstrokes-non-adversarial}
  \centering
  \includegraphics[width=0.8\linewidth]{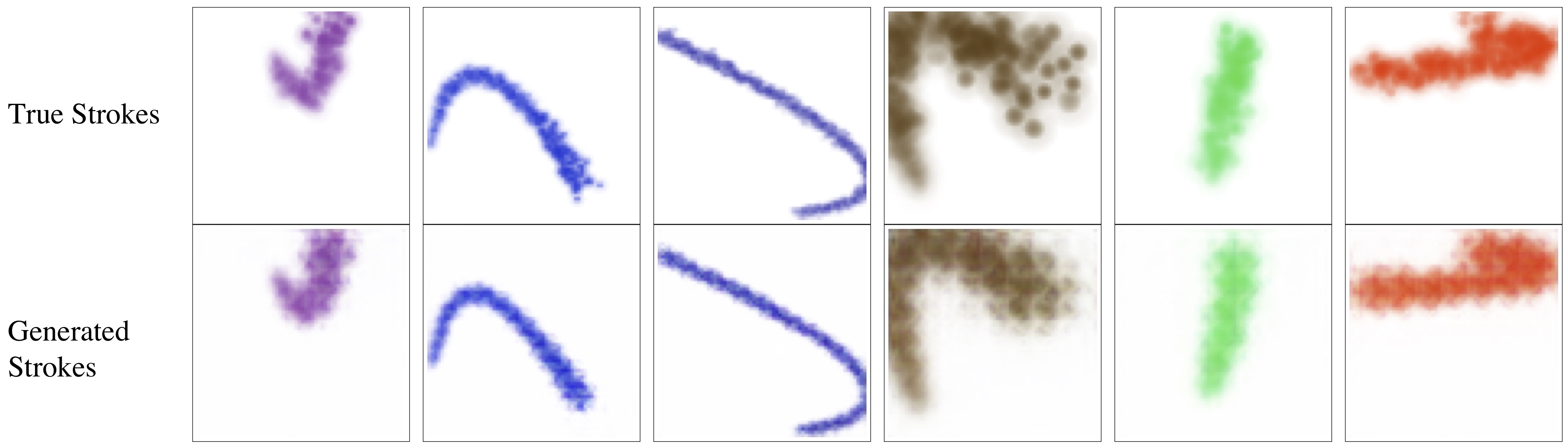}
  \caption{Sample Brushstrokes from the Generator Pre-trained with a Non-Adversarial Loss.}
\end{figure}

After pre-training the Generator using the non-adversarial loss, the brushstrokes look like the ones depicted in Figure~\ref{fig:sample-brushstrokes-non-adversarial}. A set of brushstrokes images is generated that will help us pre-train the Critic in the next step.

\subsection{Pre-train the Critic as a Binary Classifier}
\begin{figure}[ht!]
  \label{fig:neural_painter_critic_non_adversarial}
  \centering
  \includegraphics[width=0.5\linewidth]{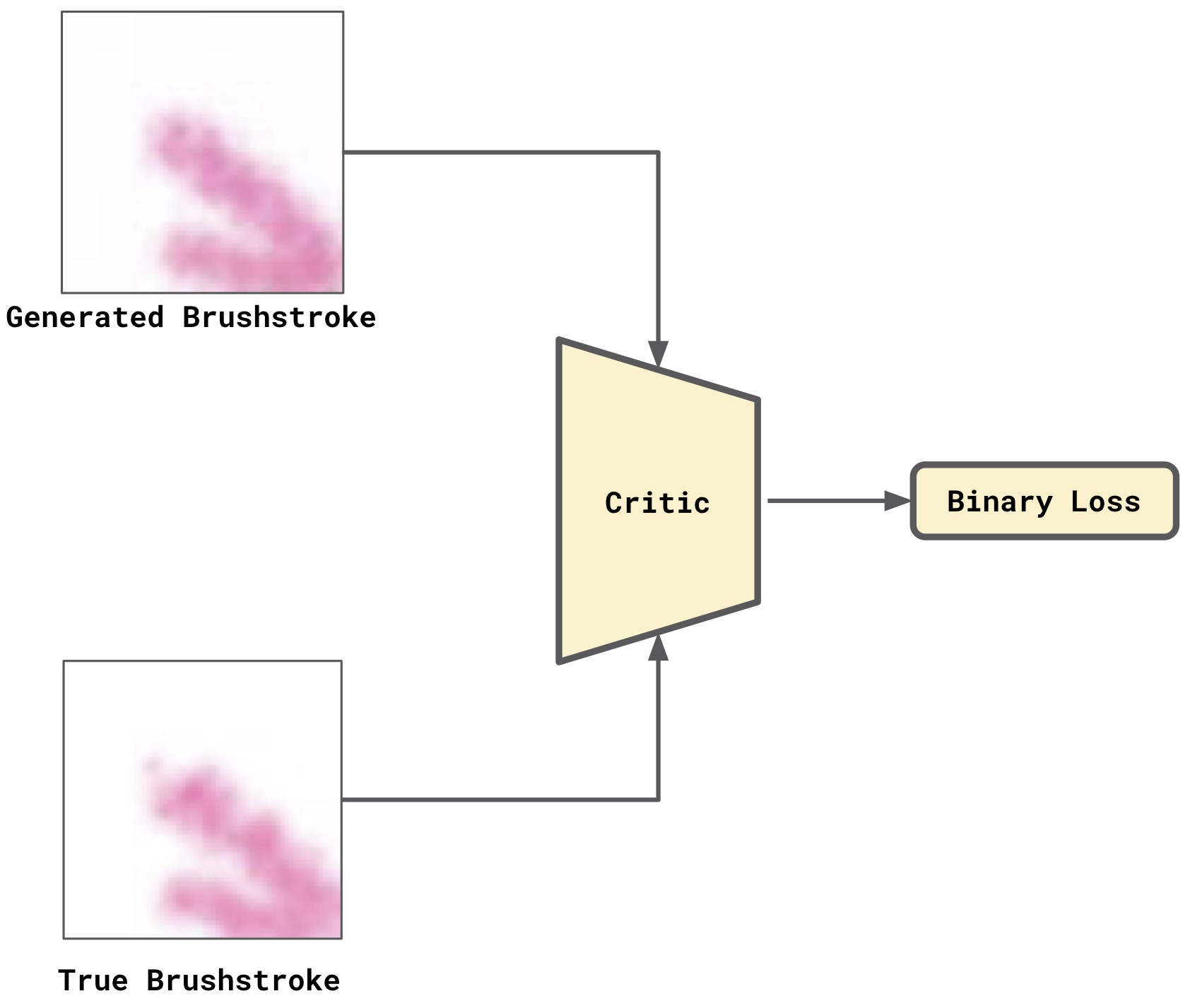}
  \caption{Pre-train the Critic as a Binary Classifier.}
\end{figure}

We train the Critic as binary classifier (Figure~\ref{fig:neural_painter_critic_non_adversarial}), that is, the Critic is pre-trained on the task of recognizing true vs generated brushstrokes images (Step (2)).
We use is the Binary Cross Entropy as binary loss for this step.

\subsection{Transfer Learning for Adversarial Training (GAN mode)}
\begin{figure}[ht!]
  \label{fig:neural_painter_generator_adversarial}
  \centering
  \includegraphics[width=0.7\linewidth]{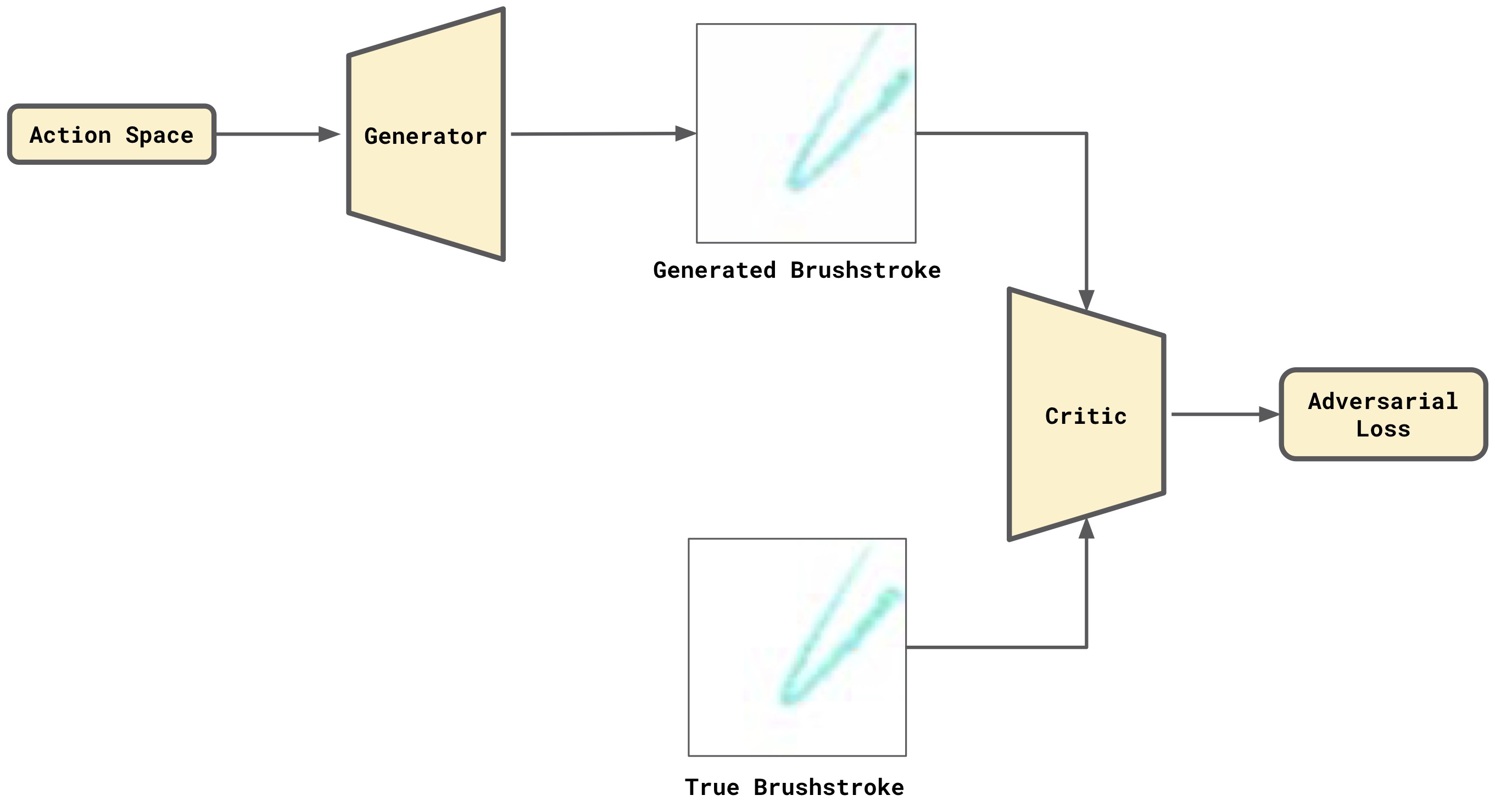}
  \caption{Transfer Learning: Continue the Generator and Critic training in a GAN setting. Faster.}
\end{figure}

Finally, we continue the Generator and Critic training in a GAN setting as shown in Figure~\ref{fig:neural_painter_generator_adversarial}. This final step is much faster that training the Generator and Critic from scratch as a GAN.

One can observe from Figure~\ref{fig:sample-brushstrokes-non-adversarial} that the pre-trained Generator is doing a decent job learning brushstrokes. However, there are still certain imperfections when compared to the true strokes in the dataset.

\begin{figure}[ht!]
  \label{fig:sample_brushstrokes_gan}
  \centering
  \includegraphics[width=0.9\linewidth]{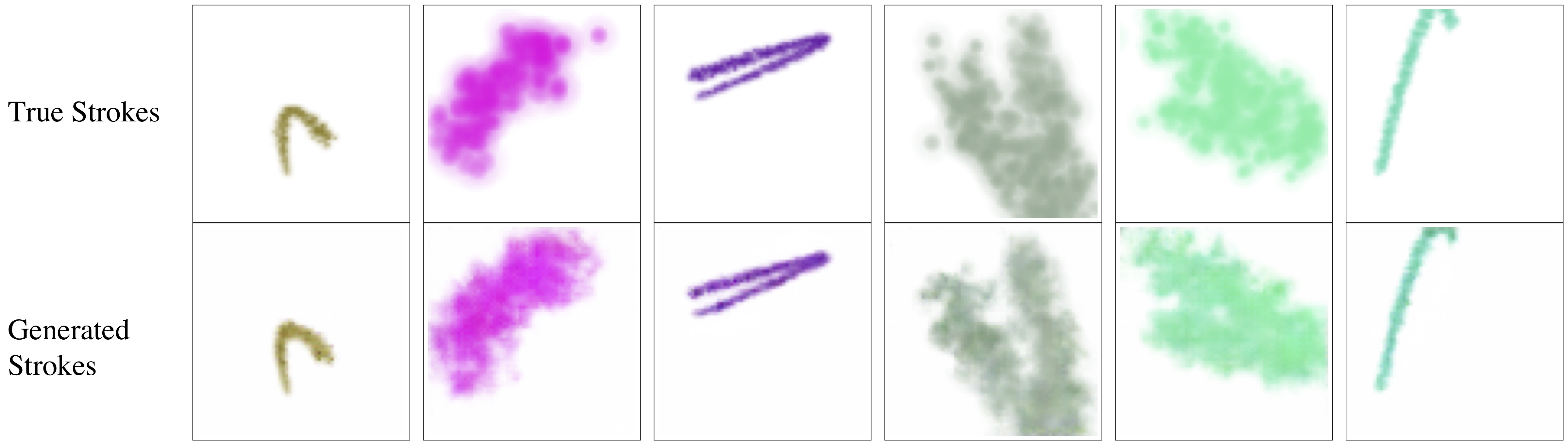}
  \caption{Sample Brushstrokes from the Generator after Adversarial Training (GAN mode).}
\end{figure}

Figure~\ref{fig:sample_brushstrokes_gan} shows the output of the Generator after completing a single epoch of GAN training, i.e., after transferring the knowledge acquired in the pre-training phase. We can observe how the brushstrokes are more refined and, although slightly different to the true brushstrokes, they have interesting textures, which makes them very appealing for brushstrokes paintings.

\subsection{Painting Generation using Brushstrokes}
Once the Generator training process is completed, we have a machine that is able to translate vectors of actions to brushstrokes, but how do we teach the machine to paint like a Bob Ross’ apprentice?

To achieve this, we use Intrinsic Style Transfer~\cite{2019NeuralPainters}, similar in spirit to Neural Style Transfer~\cite{gatys2015neural} but which does not require a style image. Intuitively, the features of the content input image and the one produced by the Neural Painter should be similar.

\begin{figure}[ht!]
  \label{fig:neural_painter_painting}
  \centering
  \includegraphics[width=0.9\linewidth]{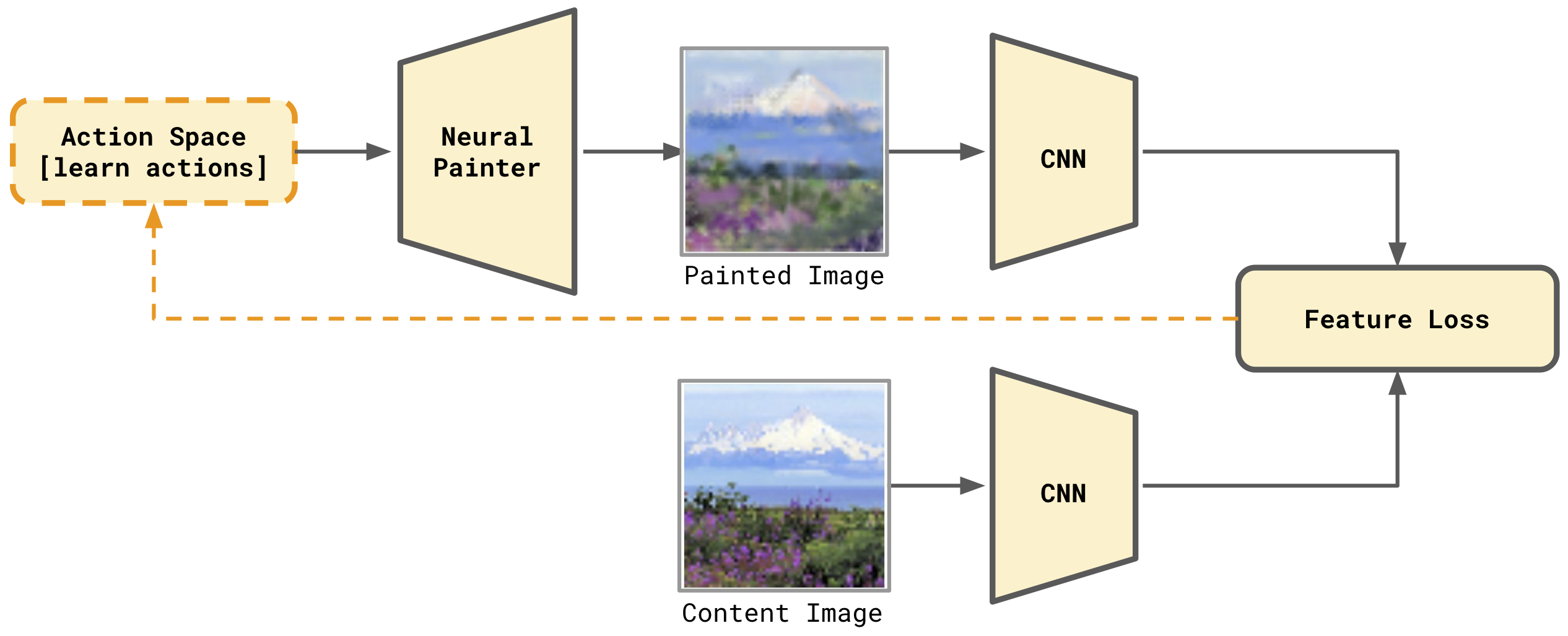}
  \caption{Painting with Neural Painters using Intrinsic Style Transfer.}
\end{figure}

To implement the process we freeze the Generator model weights and learn a set of action vectors that when input to the Generator will produce brushstrokes, that once combined, will create a painting given an input content image.The image features are extracted using a VGG16~\cite{simonyan2015deep} network as a feature extractor, denoted as CNN in Figure~\ref{fig:neural_painter_painting}, which depicts the whole process.

Note that the optimization process is targeted to learn the tensor of actions, while the remaining model weights are not changed, that is, the ones of the Neural Painter and CNN models.

\begin{figure}[ht!]
  \label{fig:painting}
  \centering
  \includegraphics[width=0.9\linewidth]{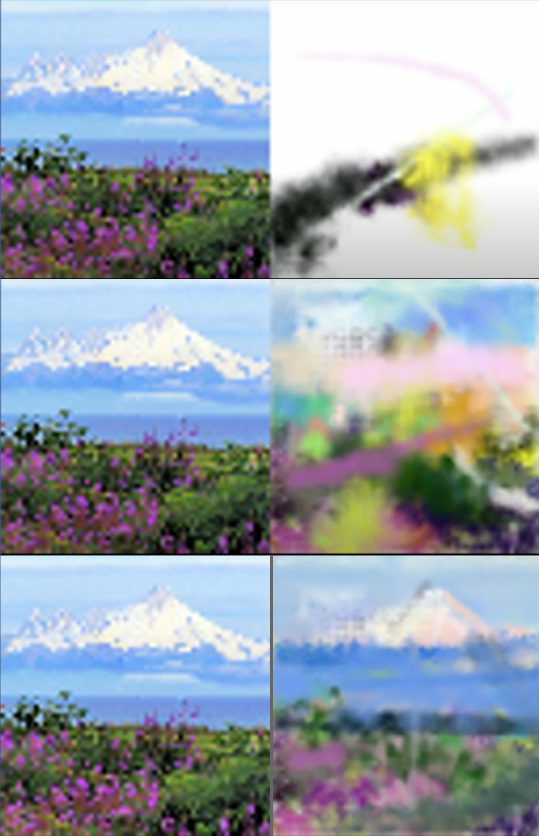}
  \caption{Progressive steps of Neural Painting generating brushstrokes to optimize a the perceptual loss of the image at the left.}
\end{figure}

\begin{figure}[ht!]
  \label{fig:pandax_collection}
  \centering
  \includegraphics[width=\linewidth]{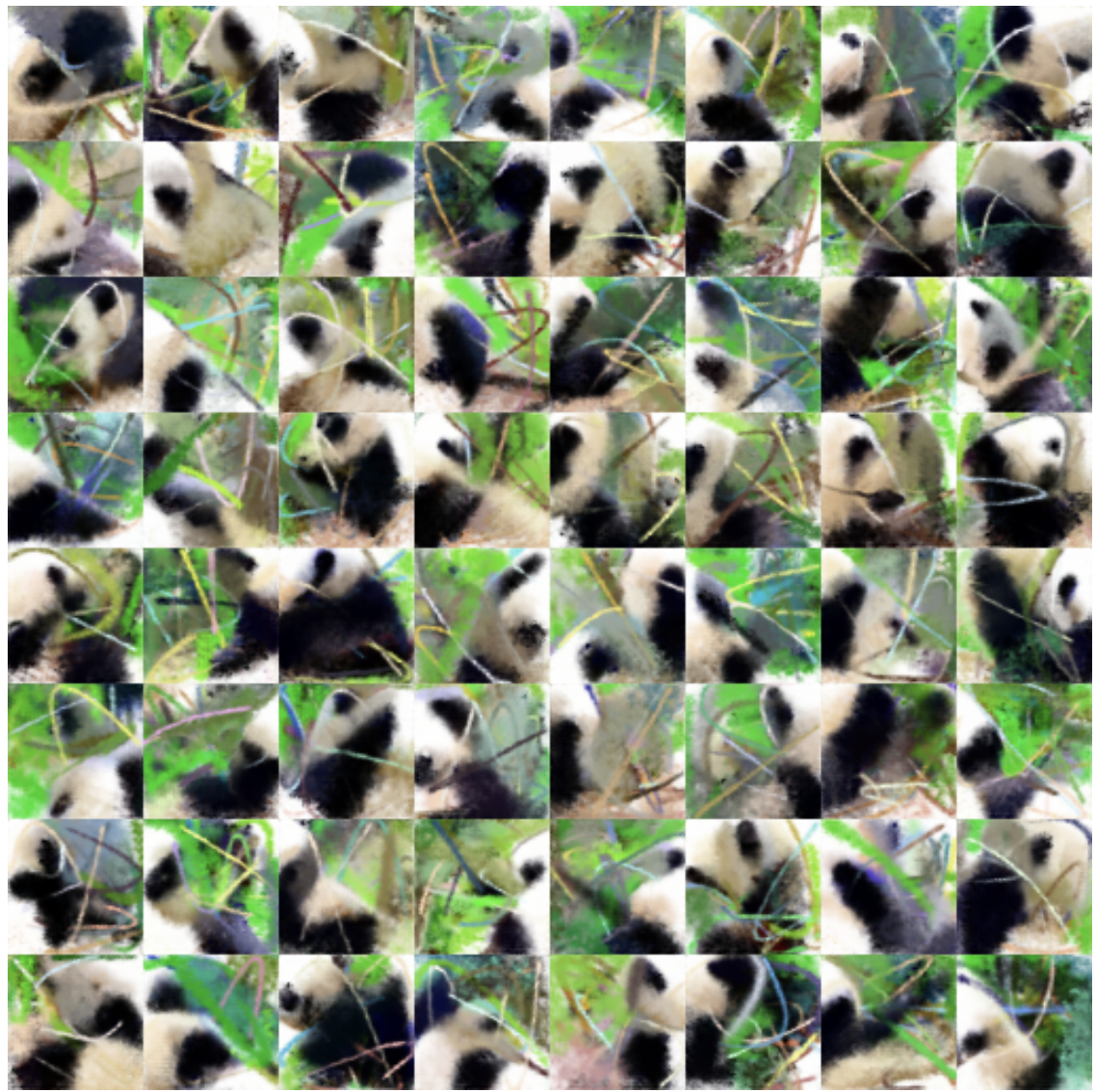}
  \caption{Neural Painter painting Pandas.}
\end{figure}

\begin{figure}[ht!]
  \label{fig:hippo_collection}
  \centering
  \includegraphics[width=\linewidth]{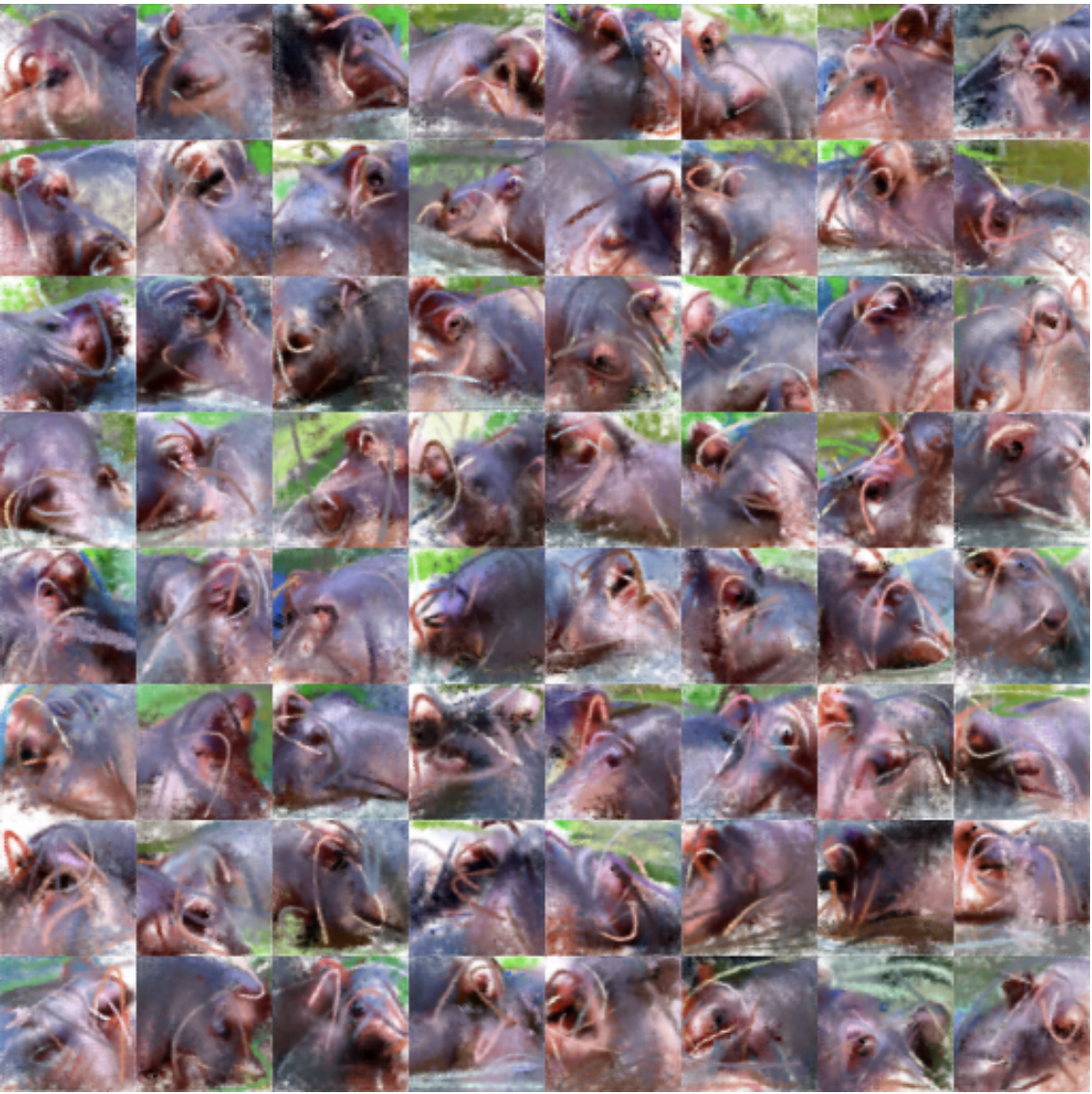}
  \caption{Neural Painter painting Hippos.}
\end{figure}

\begin{figure}[ht!]
  \label{fig:jaguars_collection}
  \centering
  \includegraphics[width=\linewidth]{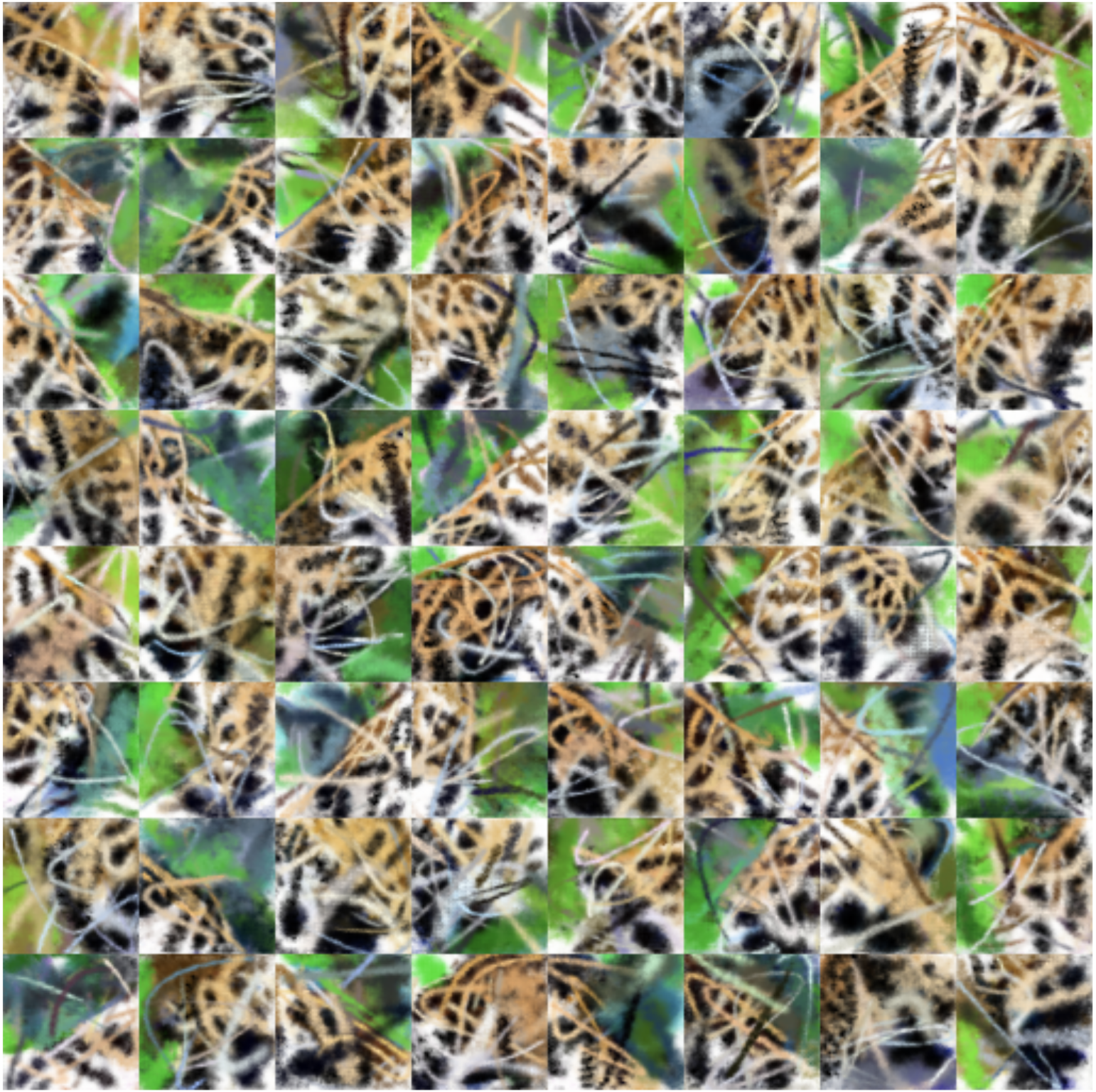}
  \caption{Neural Painter painting Jaguars.}
\end{figure}

\begin{figure}[ht!]
  \label{fig:tick_collection}
  \centering
  \includegraphics[width=\linewidth]{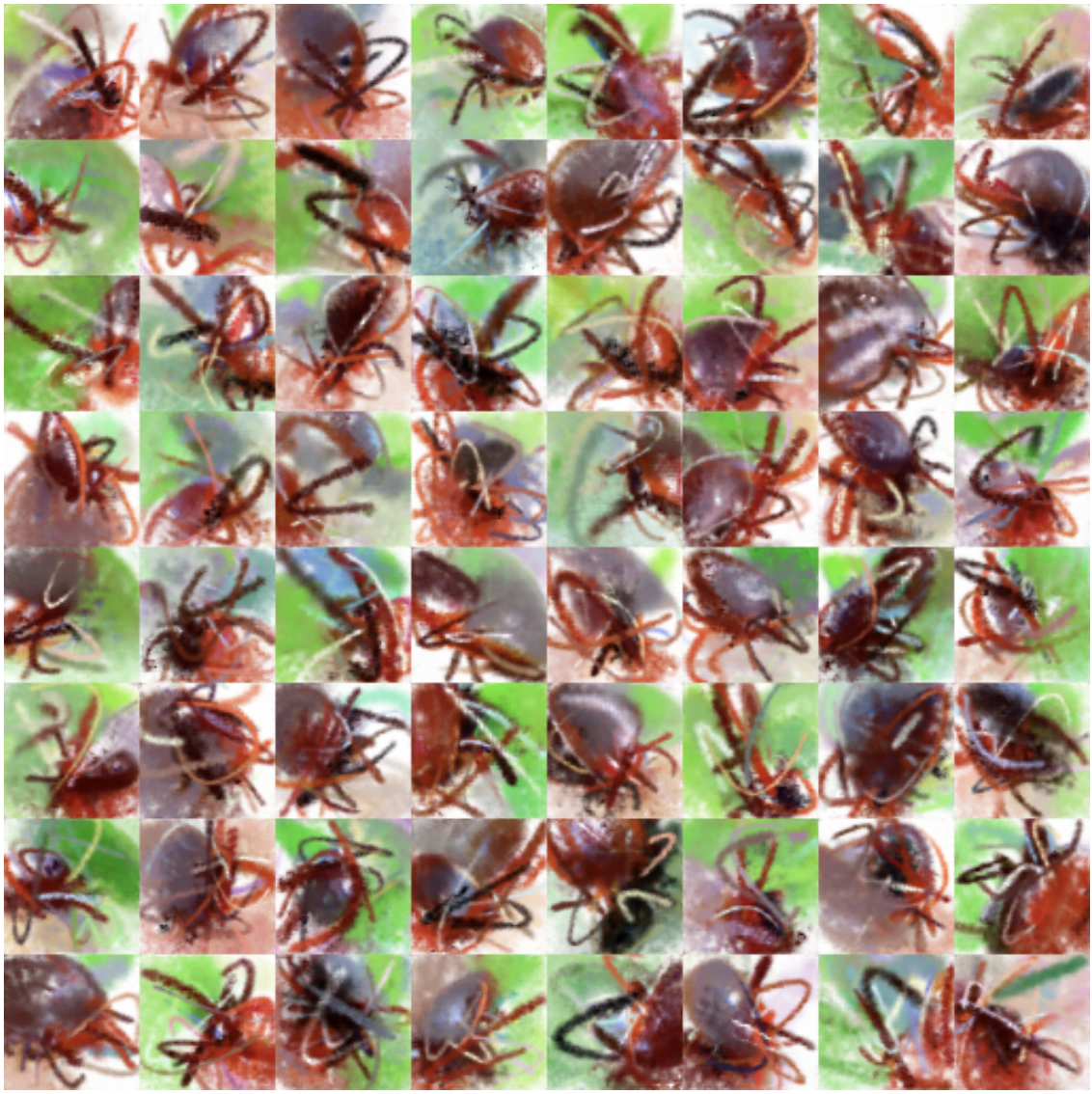}
  \caption{Neural Painter painting Ticks.}
\end{figure}

\begin{figure}[ht!]
  \label{fig:loggerhead_turtle_collection}
  \centering
  \includegraphics[width=\linewidth]{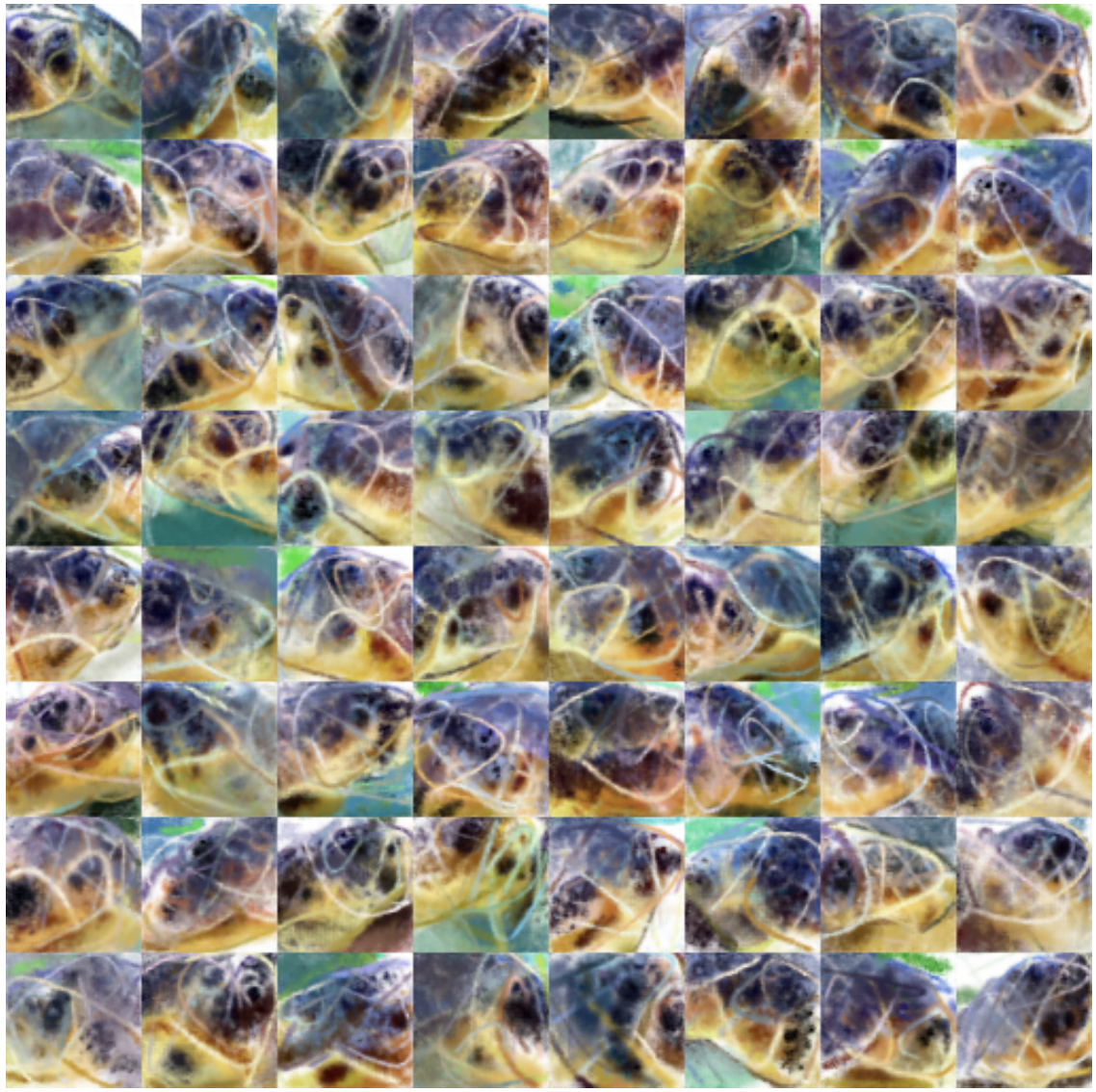}
  \caption{Neural Painter painting Loggerhead Turtles.}
\end{figure}

\begin{figure}[ht!]
  \label{fig:leatherback_turtle_collection}
  \centering
  \includegraphics[width=\linewidth]{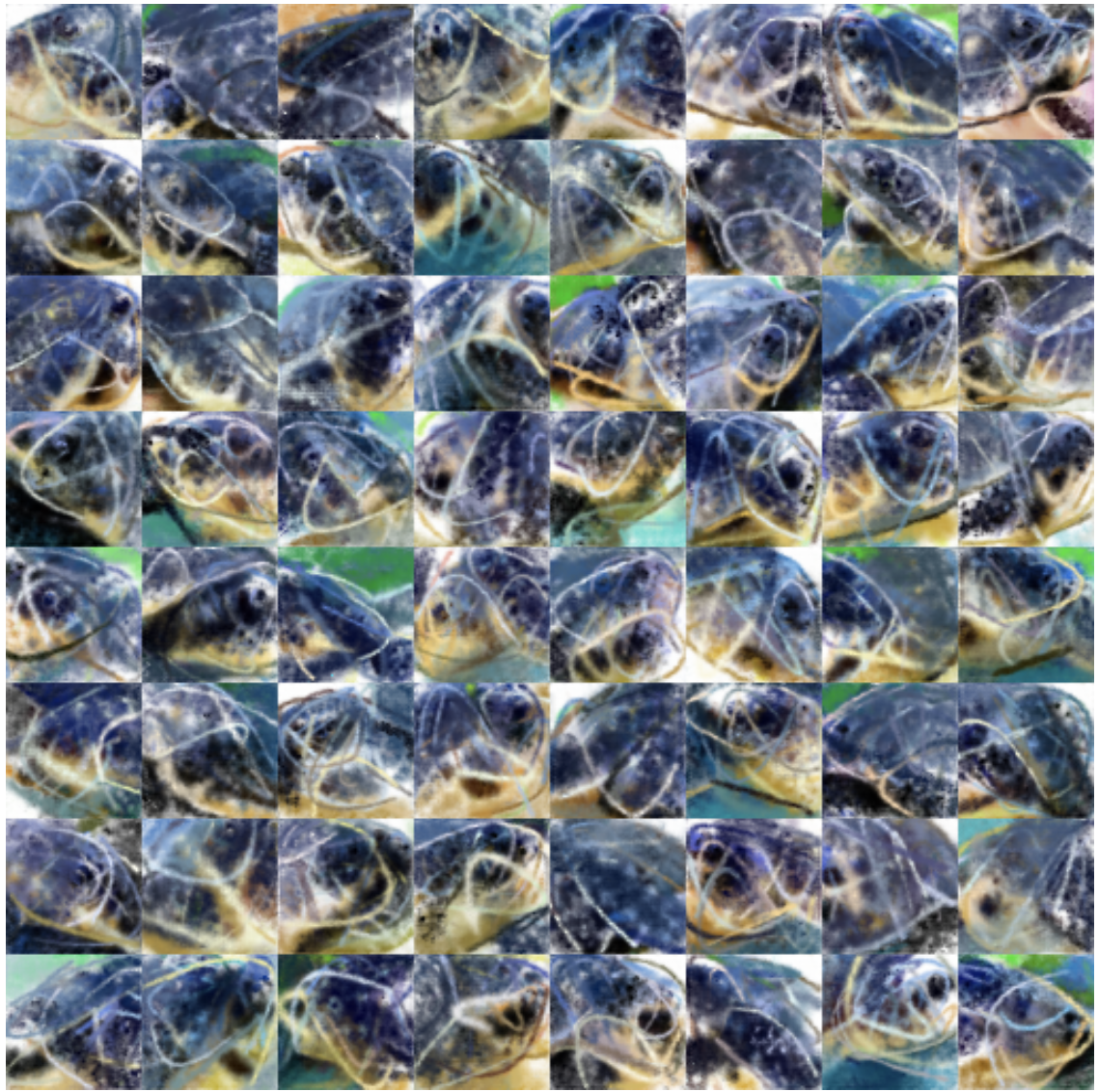}
  \caption{Neural Painter painting Leatherback Turtles.}
\end{figure}

\begin{figure}[ht!]
  \label{fig:mud_turtle_collection}
  \centering
  \includegraphics[width=\linewidth]{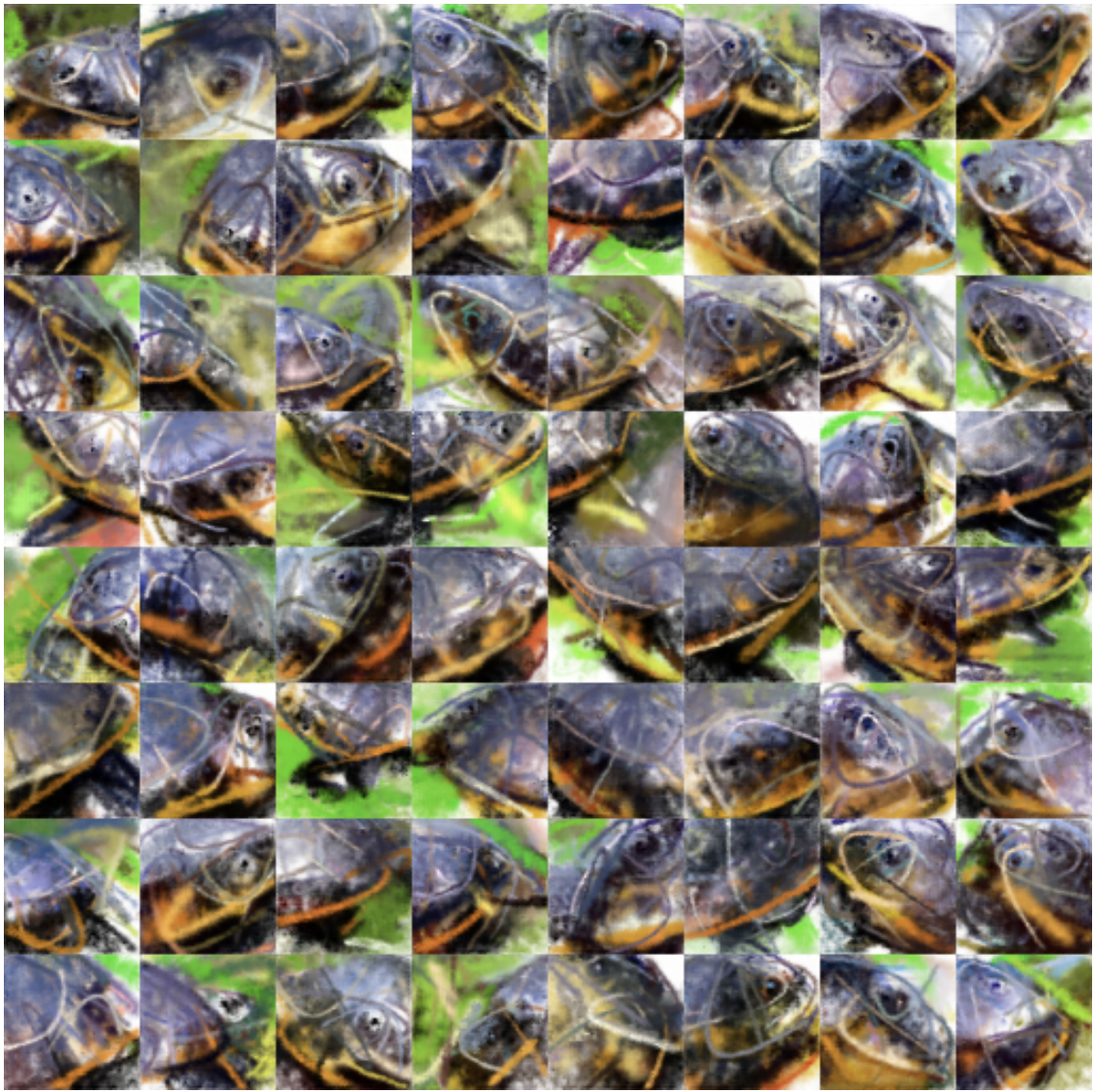}
  \caption{Neural Painter painting Mud Turtles.}
\end{figure}

\begin{figure}[ht!]
  \label{fig:owl_collection}
  \centering
  \includegraphics[width=\linewidth]{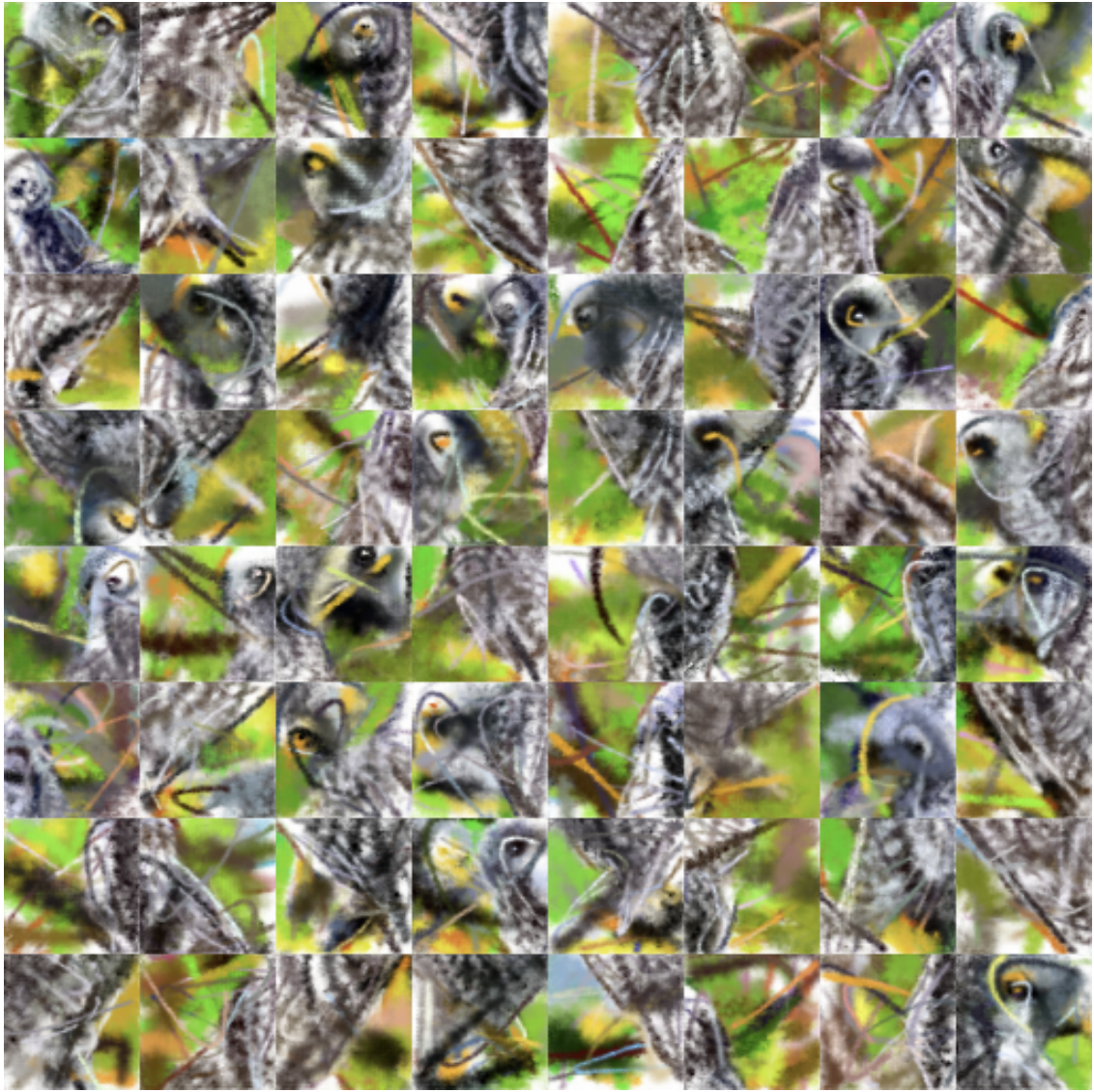}
  \caption{Neural Painter painting Loggerhead Owls.}
\end{figure}

\end{document}